\documentclass{article}

\usepackage{arxiv}

\usepackage[utf8]{inputenc} 
\usepackage[T1]{fontenc}    
\usepackage{hyperref}       
\usepackage{url}            
\usepackage{booktabs}       
\usepackage{amsfonts}       
\usepackage{nicefrac}       
\usepackage{microtype}      
\usepackage{lipsum}		
\usepackage{graphicx}
\usepackage{doi}

\usepackage{amssymb,amsthm,amsmath}
\usepackage{xcolor,paralist,hyperref,titlesec,fancyhdr,etoolbox}
\usepackage[utf8]{inputenc} 
\usepackage[T1]{fontenc}    
\usepackage{hyperref}       
\usepackage{url}            
\usepackage{booktabs}       
\usepackage{amsfonts}       
\usepackage{nicefrac}       
\usepackage{microtype}      
\usepackage{lipsum}
\usepackage{fancyhdr}       
\usepackage{graphicx}       
\graphicspath{{media/}}     

\usepackage{times}  
\usepackage{helvet}  
\usepackage{courier}  
\usepackage{graphicx} 
\usepackage{bm}
\usepackage{multirow}
\usepackage{booktabs}
\usepackage{caption} 
\DeclareCaptionStyle{ruled}{labelfont=normalfont,labelsep=colon,strut=off} 
\usepackage{graphicx}
\usepackage{algorithm}
\usepackage{algorithmic}

\newcommand{\ie}{\textit{i.e.}}

\title{Visibility Enhancement for Low-light Hazy Scenarios}

\author{
  Chaoqun Zhuang \\
  Beihang University \\
  Beijing\\
   \And
  Yunfei Liu \\
  Beihang University \\
  Beijing\\
  \And
  Sijia Wen \\
  Beihang University \\
  Beijing\\
  \And
  Feng Lu* \\
  Beihang University \\
  Beijing\\
}

\renewcommand{\shorttitle}{\textit{arXiv} Template}

\hypersetup{
pdftitle={A template for the arxiv style},
pdfsubject={q-bio.NC, q-bio.QM},
pdfauthor={David S.~Hippocampus, Elias D.~Striatum},
pdfkeywords={First keyword, Second keyword, More},
}

\begin{document}
\maketitle

\begin{abstract}
	Low-light hazy scenes commonly appear at dusk and early morning. The visual enhancement for low-light hazy images is an ill-posed problem. Even though numerous methods have been proposed for image dehazing and low-light enhancement respectively, simply integrating them cannot deliver pleasing results for this particular task. In this paper, we present a novel method to enhance visibility for low-light hazy scenarios. To handle this challenging task, we propose two key techniques, namely cross-consistency dehazing-enhancement framework and physically based simulation for low-light hazy dataset. Specifically, the framework is designed for enhancing visibility of the input image via fully utilizing the clues from different sub-tasks. The simulation is designed for generating the dataset with ground-truths by the proposed low-light hazy imaging model. The extensive experimental results show that the proposed method outperforms the SOTA solutions on different metrics including SSIM$\uparrow$ (9.19\%) and PSNR$\uparrow$ (5.03\%). In addition, we conduct a user study on real images to demonstrate the effectiveness and necessity of the proposed method by human visual perception.
\end{abstract}


\section{Introduction}

Haze is a common atmosphere phenomenon, which can be an obstacle for many computer vision tasks and multi-media applications. Though research has been proposed to improve the image quality of such phenomenon, haze removal for low-light scenarios is still a challenging problem. The hazy images captured under low-light illumination especially suffer from poor visibility, reduced contrasts, fainted surfaces, color shift, and tremendous noise. In addition, images with low-light conditions are often unavoidable for multiple computer vision tasks, such as video surveillance, autonomous driving,  urban mapping, etc. In this regard, visibility enhancement for low-light hazy scenes is indispensable for different visual-based applications.

	Conventional dehazing methods usually rely on the atmospheric scattering model~\cite{narasimhan2002vision}, which is defined as follows: 
	\begin{equation} \label{eq:normal-haze}
	I(x) = J(x)t(x) + A(1-t(x)),
	\end{equation}
	where $J$ denotes a haze-free scene radiance, $A$ describes the global atmospheric light indicating the intensity of ambient light, $t$ is the transmission map, and $x$ represents the pixel position. It notes that the decomposition of this formulation depends on the $A$ and $t$, which leads to coarse and inaccurate results under low-light illumination. As a result, the general dehazing methods cannot perform well for image dehazing under low-light illumination. 
	Similarly, the former image enhancement methods are often based on the classic Retinex theory models~\cite{wei2018deep}. The physical imaging model under low-light illumination can be expressed as: 
	\begin{equation}\label{eq:normal-enhance}
	J(x) = R(x)L(x) + \mathcal{N},
	\end{equation}
	where $R$ is the reflectance, $L$ is illumination and $\mathcal{N}$ is the illumination-related noise caused by camera intrinsic sensors. In this regard, the visibility enhancement for low-light hazy scenes can be considered as an entangled problem, which is hard to decouple.

\begin{figure} [t]
		\begin{center}
			\includegraphics[width=1.0\linewidth]{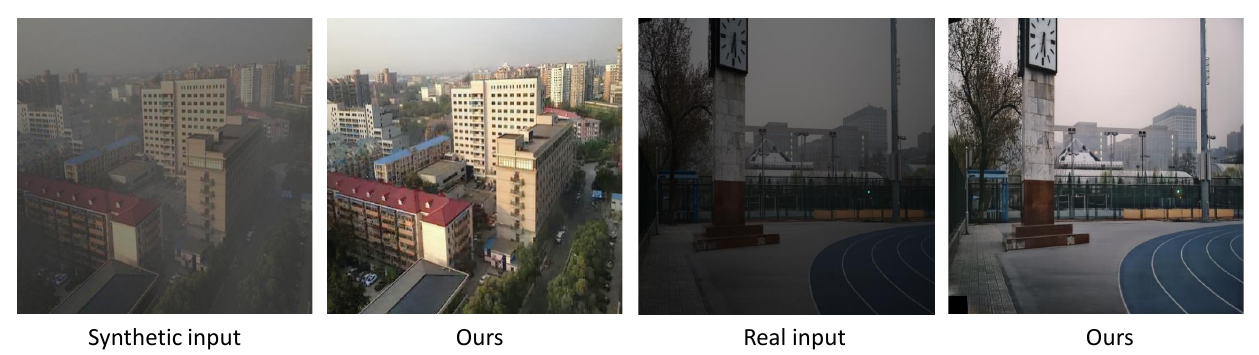} 
		\end{center}
		\caption{The proposed method can deliver pleasing visibility enhancement result for low-light hazy scene.} 
		\label{fig:teaser}
		\vspace{-5mm}
	\end{figure}

	In this paper, we propose a cross-consistency dehazing-enhancement framework to solve the aforementioned problems. The proposed method focuses on integrating dehazing inference and enhancement inference for low-light hazy scenes simultaneously. Concretely, we apply the multi-level attention-guided mappings to obtain an implicit dehazing / enhancement inference. To jointly utilize these learned mappings, we design a cross-consistency framework to integrate these inferences without affecting each other. As illustrated in Fig.~\ref{fig:teaser}, the haze is removed properly and the result images are enhanced with high visibility.
	
	Since the paired training data for low-light hazy scenes is impractical to collect, we further propose a synthetic visibility enhancement dataset. In order to simulate the visually realistic and physically reliable dataset, we make fine-tuning on the integration of the hazy imaging model and the low-light imaging model. Moreover, we consider the noise introduced by the sensor under low-light illumination. 
    
    The contribution of this work can be summarized as the following: 1) We first explore the low-light hazy scenes and propose a visibility enhancement method for this particularly difficult task. 2) We develop a cross-consistency dehazing-enhancement framework to improve the visibility of low-light hazy scenarios. 3) We propose a physical-based simulation strategy to construct the low-light hazy dataset. The simulated visibility enhancement dataset consists of 8970 images. We also capture 200 real low-light hazy scenes for verification purpose. 4) We evaluate the proposed method on the simulated dataset and the real scenes. Both experimental results prove that our solution outperforms the state-of-the-art methods. 

\section{Related Work}

	Focusing on visibility enhancement for low-light hazy scenes, the related methods can be generally grouped into three categories: image dehazing, low-light image enhancement, and image dehazing in the dark.

	\textbf{Image dehazing. }Due to the importance of image dehazing, a large number of related researches have been provided. In the beginning, some methods~\cite{pizer1987adaptive,reza2004realization} treat image dehazing as an image quality enhancement task and improve the contrast to restore a clear image by removing the noise. 	Based on the atmospheric scattering model~\cite{narasimhan2002vision}, some methods~\cite{he2010DCP,cai2016dehazenet,li2017all, zhang2018densely,liang2021fast} perform image dehazing by complying with the physical law. The most representative work is that He et al.~\cite{he2010DCP} proposed an inventive dark channel prior to estimating the transmission map for image dehazing. 
	
	Since deep learning has made great progress in recent years, the training-based network is naturally utilized in the area of image dehazing. Cai et al.~\cite{cai2016dehazenet} propose an end-to-end training model to directly estimate the transmission map, which can improve the quality of dehazing results. To improve the effectiveness of the network, Qin et al.~\cite{qin2020ffa} propose a feature fusion attention network, which combines channel attention and pixel attention mechanism. Dong~\cite{dong2020multi} newly propose a network with dense feature fusion based on the U-Net architecture. Shao et al.~\cite{Shao_2020_CVPR_DA}, Liu et al.~\cite{liu2021syntheticMM} and Chen et al.~\cite{chen2021psdcvpr} focus on real hazy data, and try to improve the performance of dehazing on real data. Zheng et al.~\cite{zheng2021ultra} proposed a multi-guide bilateral learning framework for 4K resolution image dehazing.  Sun et al.~\cite{sun2022sadnet} explore the effect of the semi-supervised method in the field of image dehazing. There are also some researches~\cite{liang2021fast, du2021real} that try to explore real-time dehazing models.
	
	Although these methods can remove haze from general hazy images, due to the physical complexity of low-light hazy scenes, the image dehazing cannot be performed correctly in low-light environments. In addition, since the haze is highly related to the depth information of the scene, these methods cannot deliver pleasing dehazing results.

	\textbf{Low-light image enhancement. }The existent methods~\cite{pizer1987adaptive,huang2012gama_correct,wei2018deep,hao2022decoupled} try to address this issue through statistical assumptions and computational optimization. In recent years, Shen et al.~\cite{shen2017MSR} firstly develop an end-to-end multi-scale Retinex convolutional network to achieve low-light image enhancement. Lv et al.~\cite{lv2018mbllen} further propose a multi-branch low-light image enhancement network to remove the artifacts and noise in dark areas. Zhang et al.~\cite{zhang2019kindling} decompose the image into brightness and reflection parts. The former part is used to adjust the light flexibly, and the latter part is used to deal with the noise and color distortion. Guo et al.~\cite{Zero-DCE2020} convert this task into a specific curve estimation problem. Mohit et al.~\cite{lamba2021restoring} try to deal with this problem in an extremely dark scenario and achieve it at a real-time level. Yet these methods will be seriously disturbed by the haze in the dark environment, which amplifies the noise during the image enhancement.
	
	\textbf{Image dehazing in the dark. }By considering the artificial light sources at night as the ambient light, some methods develop the night dehazing model for image dehazing in the dark. These methods normally solved this problem by setting some priors~\cite{zhang2014nighttime,li2015nighttime,zhang2017fast} or simulating the night hazy images~\cite{zhang2020nighttime}. However, the expected results of these nighttime dehazing methods cannot deliver the fine details in terms of visual perception. 
	
	In summary, none of the above-mentioned methods can effectively improve the image quality of low-light hazy scenes. In this paper, we focus on not only the dehazing results but also the visibility enhancement for low-light hazy images.

	\section{Proposed Method}
	\subsection{Problem Analysis}
	Since the low-light hazy images comply with a complex physical imaging model, the decomposition of such a model is the key to tackling this ill-posed problem. To begin with, we first formulate the task with precise denotations. Specifically, we denote $I$ as the well-exposed hazy-free image. $I_L$ is the low-light image without haze. $I^H$ is the hazy image under normal exposure. $I^H_L$ is the hazy low-light image. Our goal is to learn a mapping $\mathcal{S}(\cdot, \theta)$ to recover the well-exposed haze-free image $I$ from the under-exposed hazy image $I^H_L$, where $\theta$ denotes the learnable parameters for implicit inference. 
	
	We firstly propose a low-light enhancement mapping $\mathcal{E}(\cdot, \theta)$, which converts the images captured under low-light illumination to well-exposed images with rich details. In addition, we denote $\mathcal{D}(\cdot, \theta)$ as a dehazing map for haze removal. Apparently, there are two paths to obtain the well-exposed haze-free image $I$ from $I^H_L$. 
	\begin{figure*} [t]
		\begin{center}
			\includegraphics[width=1.0\linewidth]{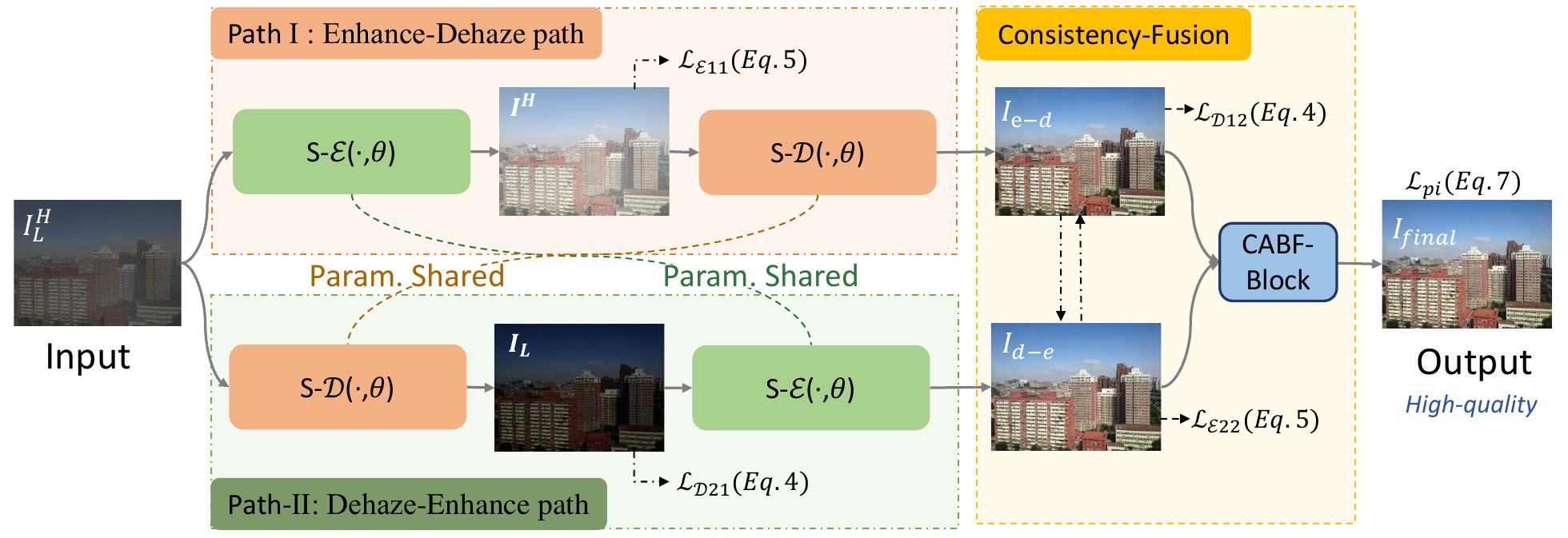} 
		\end{center}
		\caption{Overview of our cross-consistency dehazing-enhancement framework.} 
		\label{fig:overview}
	\end{figure*}
	
	\begin{enumerate} \label{two-path}
		\item $LLE \rightarrow Dehaze$. Applying the low-light enhancement mapping $\mathcal{E}$ on the low-light hazy image $I^H_L$ can deliver an intermediate result $I^H$, which should be a well-exposed image with haze residual. Then we adopt the dehazing mapping $\mathcal{D}$ for $I^H$ to obtain the final results. Formally, the forward process is $I = \mathcal{D}(I^H, \theta)$, where $I^H = \mathcal{E}(I^H_L, \theta)$.
		\item $Dehaze \rightarrow LLE$. Similarly, we operate the dehazing first, then perform low-light enhancement. In this regard, the forward process is $I = \mathcal{E}(I_L, \theta)$, where $I_L = \mathcal{D}(I^H_L, \theta)$, where the intermediate $I_L$ is the under-exposed image without haze.
	\end{enumerate}
	
	To better address this problem, we introduce an order-invariant assumption. From a physical perspective, the sequence of the two natural phenomena of low light and haze will not impact the imaging. From a mathematical perspective, the order of dehazing and enhancement theoretically does not affect the results. Therefore, we assume that the computational process of low-light enhancement and dehazing is order-invariant, which is proved by our experimental results in Table~\ref{tab:ablation_study}. Based on this assumption, we propose the mapping $\mathcal{S}$ to jointly utilize $\mathcal{E}$ and $\mathcal{D}$ with customized constraints. The computational process can be formulated as:
	\begin{equation} \label{eq:forward}
	\begin{matrix}
 \mbox{Input image\;}I^H_L \\
\textcolor[RGB]{192,0,0}{\mbox{\, Dim \& Hazy}}
\end{matrix}
	\begin{array}{l}{\nearrow} \\ \\{\searrow}\end{array}
	\begin{array}{l}{\stackrel{\mathcal{E}(\cdot, \theta)}{\longrightarrow} I^H \stackrel{\mathcal{D}(\cdot, \theta)} {\longrightarrow} I_{e-d}} \\  
\textcolor[RGB]{192,0,0}{Path\,\uppercase\expandafter{\romannumeral1}: LLE \rightarrow Dehaze} \\ \\
	{\stackrel{\mathcal{D}(\cdot, \theta)}{\longrightarrow} I_L \stackrel{\mathcal{E}(\cdot, \theta)} {\longrightarrow} I_{d-e}} \\
	\textcolor[RGB]{192,0,0}{Path\,\uppercase\expandafter{\romannumeral2}:Dehaze \rightarrow LLE} \end{array} 
	\begin{array}{l}{\searrow} \\  \\ {\nearrow}\end{array}
	\begin{matrix}
\mathcal{F}(\cdot, \theta) \\
\longrightarrow \\
\textcolor[RGB]{192,0,0}{\mbox{Fusion}}
\end{matrix}
	\begin{matrix}
\mbox{Output image\;}I_{final}\, , \\
\textcolor[RGB]{192,0,0}{\mbox{\, High-quality}}
\end{matrix}
	\end{equation}
	where $I_{e-d}$ is the output of first low-light enhancement then dehaze path, and $I_{d-e}$ is the output of first dehaze then low-light enhancement path. We then fuse the two results using a mapping $\mathcal{F}(\cdot, \theta)$ to get the final result.
	
	\subsection{Framework}
	As mentioned above, we propose a cross-consistency dehazing-enhancement framework, which consists of pre-tuned dehazing and enhancement mappings. Essentially, these two mappings need to be capable of inferring clean/well-exposed images from hazy/low-light images. For this purpose, each mapping is constructed in two parts: 1) a rich-encoder for multi-level feature extraction and 2) an attention-guided decoder.
	
	\textit{Rich-encoder}. Based on Resnet-18~\cite{he2016deep}, we extract multi-level features by gathering intermediate output from each stage of the backbone to preserve more detailed information. Then the encoder produces mixed features from different levels of the encoding process. The mixed features contain rich information, which can help for a more effective final inference. 

	\textit{Attention-guided decoder}. The decoder's purpose is to map the learned features from high-dimensional space to RGB space. Specifically, given the input features, the decoder first aggregates them by bilinear-resizing each of the features to the same size as the input image, and then channel-wisely concatenates them together. Inspired by the attention mechanism~\cite{zhang2018image,wei2019single,liuICCV2019GridDehazeNet}, we apply the channel-spatial attention~\cite{wei2019single} for the context between channels and multi-scale context within channels. 
	
    \begin{figure*} [b]
		\begin{center}
			\includegraphics[width=1.0\linewidth]{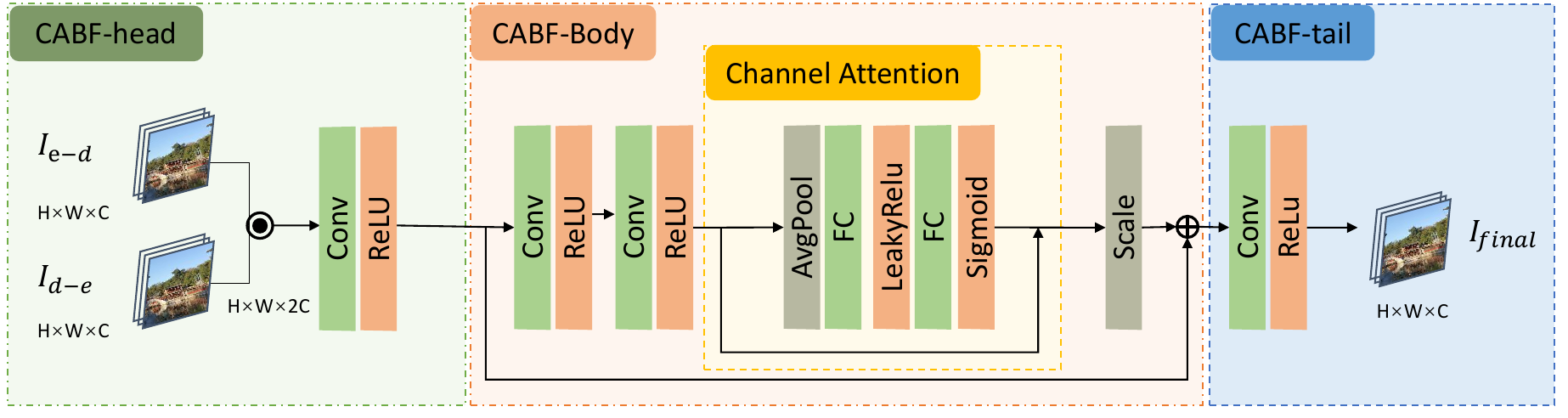} 
		\end{center}
		\caption{Architecture of the CABF (channel-attention-based fusion block). This block fuses two different path's results to get a high-quality final result.} 
		\label{fig:fusion_block}
		\vspace{-3.5mm}
	\end{figure*}

    \textit{Cross-consistency dehazing-enhancement architecture. }After obtaining these two mappings, we apply a unified architecture to jointly use these two mappings with customized constraints. An overview of the proposed method is illustrated in Fig.~\ref{fig:overview}. According to the order-invariant assumption, we want to jointly use these two mappings for performance improvement without affecting each other. To perfectly utilize the consistency of these two paths, we share the parameters of these mappings, which have the same purpose in different paths. We use $\mathcal{S-D}(\cdot, \theta)$ and $\mathcal{S-E}(\cdot, \theta)$ to represent that the two mappings share parameters in Fig.~\ref{fig:overview}. Then, we design the CABF block to fuse the results of these two paths. 

    \textit{Channel-Attention-Based Fusion block}. To achieve better image quality, we carefully design a channel-attention-based fusion block(CABF) to combine the results from two paths. The underlying design principle is to introduce channel-wise context for the final output.
	As shown in Fig.~\ref{fig:fusion_block}, the fusion block takes two images as input, \ie, $I_{e-d}$ and $I_{d-e}$. Through utilizing the imperceptible clues from two inputs, it fuses them in latent space and generates the refined image $I_{final}$.
	More specifically, the CABF block first concatenates two images in the channel dimension. Then it enhances the local receptive field via a stack of CNNs. Next, we adopt SE-layer to adaptively predict the relative importance of each channel. The overall structure is within a residual block, without increasing the parameter count. Finally, the output layer is a convolutional layer with a ReLU activation for reconstructing the refined image.
	
	Based on the proposed architecture, the dehazing mapping can infer clean images from the hazy ones under both low-light and normal-exposed conditions. Similarly, the enhancement mapping is capable of bridging the gap between low-light images and normal ones with dense haze. In other words, such shared parameters can make the mapping focus on their own task and discard the interference of other irrelevant factors. Our framework implements logical two-path processing using two basic mappings and a fusion block.

	\noindent\textbf{Loss functions.}
	We apply different loss functions for these two types of mappings. For Dehazing mapping, the loss function is 
	\begin{equation} \label{eq:haze_loss}
	\mathcal{L}_\mathcal{D}(\mathbf{x}, \hat{\mathbf{x}}) = |\mathbf{x} - \hat{\mathbf{x}}|_1 + |\mathbf{x} - \hat{\mathbf{x}}|_2 + |\nabla \mathbf{x} - \nabla \hat{\mathbf{x}}|_1,
	\end{equation}
	where $\hat{\mathbf{x}}$ and $\mathbf{x}$ are prediction and ground-truth, correspondingly. $\nabla$ denotes the gradient operation, which is used as a constraint for the edge of the output.
	
	For Enhancement mapping, the loss function is 
	\begin{equation} \label{eq:enhancement_loss}
	\mathcal{L}_\mathcal{E}(\mathbf{x}, \hat{\mathbf{x}}) = |\mathbf{x} - \hat{\mathbf{x}}|_1 + |\mathbf{x} - \hat{\mathbf{x}}|_2 + \mathcal{L}_{exp}(\hat{\mathbf{x}}),
	\end{equation}
	where $L_{exp}(\cdot)$ is a self-supervised loss for controlling the exposure level~\cite{mertens2007exposure}. This loss leverages the distance between the average intensity value of a local region to a well-exposed level $\delta$. We use the same settings as the existing work~\cite{Zero-DCE2020} in our experiment. The $L_{exp}$ is expressed as:
	\begin{equation}
	\mathcal{L}_{exp} = \frac{1}{M}\sum^M_{k=1}|y_k - \delta|,
	\end{equation}
	where $M$ denotes the number of non-overlapping patches of 16$\times$16, $y_k$ is the average value of k-\textit{th} patch in the predicted image $\hat{\mathbf{x}}$. 
	
	Moreover, we add a path-invariant loss $L_{pi}$ to keep the outputs from two paths consistent, \ie,
	\begin{equation}
	\mathcal{L}_{pi} = |\mathcal{E}(\mathcal{D}(I_L^H)) - \mathcal{D}(\mathcal{E}(I_L^H))|.
	\end{equation}
	
	Finally, the total loss of our proposed method is 
	\begin{equation} \label{eq:total-loss}
	\mathcal{L}_{total} = \lambda_{11} \mathcal{L}_{\mathcal{E}11} + \lambda_{12} \mathcal{L}_{\mathcal{D}12} + \lambda_{21} \mathcal{L}_{\mathcal{D}21} + \lambda_{22} \mathcal{L}_{\mathcal{E}22} + \lambda_3 \mathcal{L}_{pi},
	\end{equation}
	where $\lambda_{11}$, $\lambda_{12}$, $\lambda_{21}$, $\lambda_{22}$ and $\lambda_3$ are tunable parameters that make the total loss flexible. 
	
    
    

	\section{Simulation and Dataset}

	\subsection{Low-light Haze Simulation}
	\begin{figure} [t]
		\begin{center}
			\includegraphics[width=0.9\linewidth]{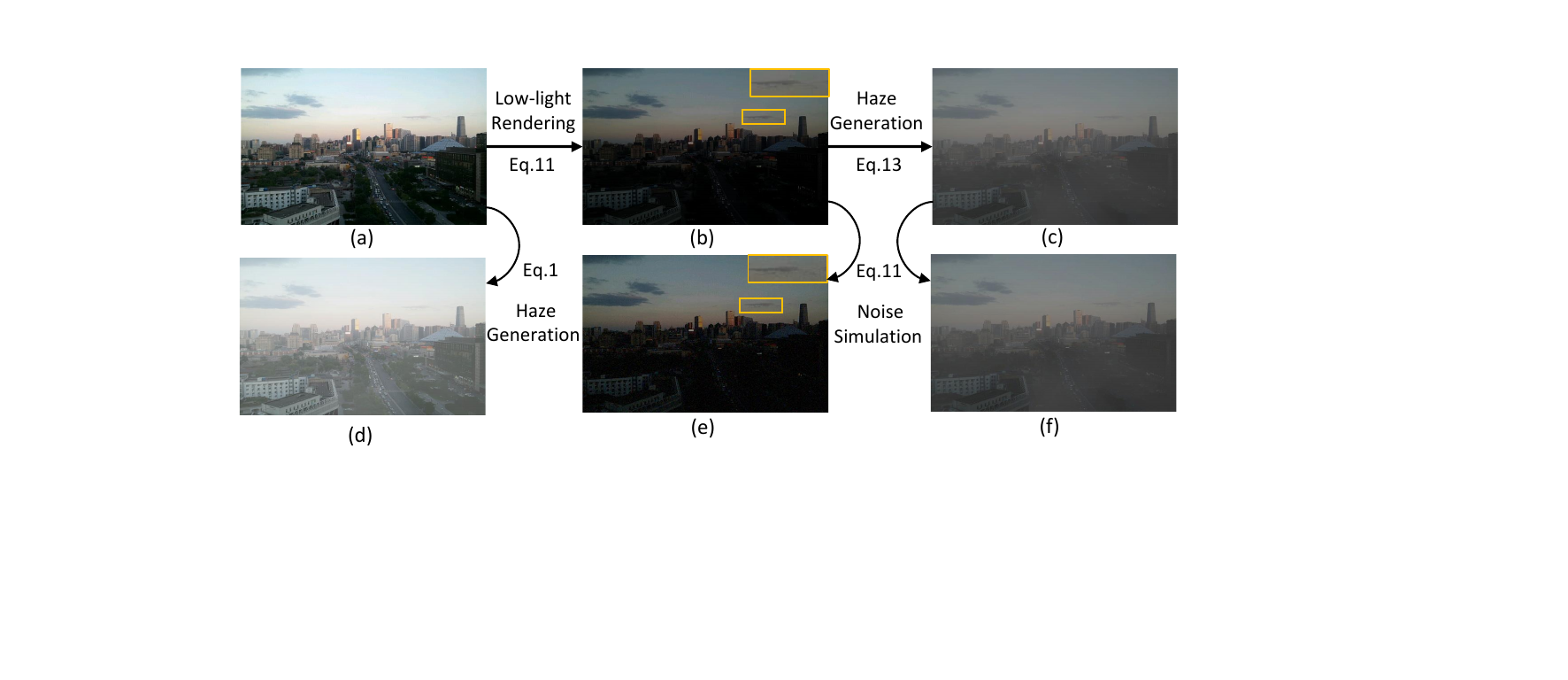} 
		\end{center}
		\caption{Pipeline of our data simulation. (a) Clear image. (b) Simulated low-light image. (c) Simulated low-light hazy image. (d) Simulated haze-only image. (e) Simulated low-light-only image with noise. (f) Our final low-light hazy simulation.} 
		\label{fig:data_construction}
		\vspace{-3mm}
	\end{figure}
	
	It should be noted that suitable datasets for our motivation do not exist. Moreover, there is no long-range sensor to directly collect the visibility enhancement training pair. Therefore, we propose a physical-based simulation strategy to collect the first visibility enhancement dataset for low-light hazy scenes. As illustrated in Eq.~\ref{eq:normal-haze}, the global atmospheric light $A$ relies on the intensity of the illumination. Therefore, to simulate the haze under low-light illumination, we should render the low-light environment first, then simulate the haze. The pipeline of the simulation can be generally summarized into three phases: 1) low-light rendering; 2) haze generation; 3) noise simulation, as shown in Fig.~\ref{fig:data_construction}.
	
	\textit{Low-light rendering. }To render the low-light haze in terms of visual perception and physical constraints, we utilize the Retinex theory~\cite{land1977retinex} for the low-light imaging model: 
	\begin{equation}
	I(x)   = R(x) L(x), \quad
	I_{low}(x) = R(x) L_{low}(x) + \mathcal{N},
	\end{equation}
	where $R$ stands for reflection, $L$ stands for illumination. $\mathcal{N}$ is the unbalanced illumination distribution and the noise caused by camera intrinsic sensors. Afterwards, we can get the relationship between normal image $I$ and the corresponding low-light image $I_{low}$ as following:
	\begin{equation} \label{eq:low-light-imagery}
	I_{low}(x) = I(x) \frac{L_{low}(x)}{L(x)} + \mathcal{N}
	= I(x) \Delta L(x) + \mathcal{N},
	\end{equation}
	where $\Delta L(x)$ represents the various lightness on objects caused by the low-light illumination. In conclusion, $\Delta L(x)$ is the key parameter to render the low-light imagery.
	
	Since just setting a prior value for $\Delta L(x)$ cannot simulate the complex scenarios with low-light illumination, we introduce auxiliary variables to reformulate this delicate parameter. Based on the analysis of the imaging model with different exposure levels and other low-light image synthetic methods~\cite{lore2017llnet,lv2020fast}, we use a combination of linear adjustment and gamma adjustment to fit the function of $\Delta L(x)$. In this case, Eq.~\ref{eq:low-light-imagery} can be reformulated as:
	\begin{equation}\label{low-light-adjust}
	I_{low}(x) = \beta \ (\alpha \  I(x))^{\gamma} + \mathcal{N},
	\end{equation}
	where $\alpha$, $\beta$ and $\gamma$ are parameters for adjusting exposure levels to simulate the under-exposed image. After numerous analytical tests on real scenes with different levels of exposure, we randomly set these three parameters from the certain value ranges: $\alpha \in [0.9, 1]$, $\beta  \in [0.5, 0.7]$ and $\gamma \in [1.5, 2.5]$.
	
	\textit{Haze generation. }After obtaining the low-light data, inspired by RESIDE~\cite{li2017reside}, we can generate the haze as following:
	\begin{equation} \label{eq:haze-imagery}
	I^{hazy}_{low}(x) = I_{low}(x)t(x) + A (1 - t(x)),
	\end{equation}
	where $I^{hazy}_{low}(x)$ denotes a low-light hazy image. By plugging Eq.~\ref{eq:low-light-imagery} into Eq.~\ref{eq:haze-imagery}, the low-light hazy image can be generated through:
	\begin{equation} \label{eq:haze-imagery1}
	I^{hazy}_{low}(x) = \beta \ (\alpha \ I(x))^{\gamma}t(x) + A (1 - t(x)) + \mathcal{N}.
	\end{equation}
	We select the brightest pixels in $I_{low}(x)$ to estimate $A$. Based on the depth of RESIDE, the scattering coefficient of $t$ is randomly set within [0.1, 0.2].

	\textit{Noise simulation. }As for noise, we follow a realistic noise model ~\cite{Guo2019Cbdnet}, which considers the influence of signal-dependent noise and the ISP process. 
	
	Mathematically, we can obtain the low-light hazy image with an arbitrary mutually combination of Eq.~\ref{eq:low-light-imagery} and Eq.~\ref{eq:haze-imagery}. As we explained before, the global atmospheric light $A$ and the noise $\mathcal{N}$ are highly related to the intensity of illumination in real scenes. Therefore, it is reasonable to render the low-light environment before generating the haze.

	\subsection{Datasets}
	  
	Based on the RESIDE dataset, we simulate 8970 paired low-light hazy scenes by the above-mentioned simulation strategy. We split the simulated dataset into 8073 training images and 897 testing images. To constrain the dehazing / enhancement block during the training phase, we also generate the haze-only dataset and low-light-only dataset. In conclusion, the proposed simulated dataset includes four groups, which is formulated as $X = \{I,I_L,I^H, I^H_L\}$.
	
	Since there is no real data for this newly raised task, we also collect a real low-light hazy dataset, which contains 200 images. These images were taken with a Sonny $\alpha$7R\uppercase\expandafter{\romannumeral2} camera on a tripod, which are as following considerations: scenes (\ie houses, campuses, streets), camera viewing angles (\ie front view and oblique view), camera apertures (\ie f/2.0 - f/16).  The dataset and corresponding project will be released soon.

	\section{Experiments}
	
	Since our task is for low-light hazy images, which involves two tasks: dehazing and low-light enhancement. We choose the SOTA dehazing and low-light enhancement methods for comparison experiments: FFA~\cite{qin2020ffa}, MSBDN~\cite{dong2020multi}, DA~\cite{Shao_2020_CVPR_DA}, Grid~\cite{liuICCV2019GridDehazeNet}, KinD~\cite{zhang2019kindling}, and ZeroDCE~\cite{Zero-DCE2020}. For fair comparisons, these methods are evaluated on the proposed simulated dataset with retrained models. We evaluate the quantitative comparison by the commonly used metrics: PSNR and SSIM~\cite{wang2004image}. It should be noted that the newly raised task is about visibility enhancement for low-light hazy scenes, and there is no specific method to directly solve this problem. In this regard, to demonstrate that the simple sequential combination of dehazing and enhancement methods cannot solve this particular task, we utilize two strategies for comparison:  $LLE \rightarrow Dehaze$ (First low-light enhancement and then dehaze strategy) and $Dehaze \rightarrow LLE$ (First dehaze and then low-light enhancement strategy) methods. In addition, we directly retrain the above-mentioned methods to prove the effectiveness of our network. 
	
	In this section, firstly, we introduce our experimental implementation details. Next, we conduct a series of comparisons among our methods and other SOTA approaches on our synthetic visibility enhancement dataset, including two different sequential combination ways and the end-to-end way. Then we show the performance of our method in real low-light hazy scenarios. Furthermore, we conduct ablation experiments to demonstrate the effectiveness of our framework. Finally, we do some applied research to discuss the value of our method.
	\begin{figure*} [t]
		\begin{center}
			\includegraphics[width=\linewidth]{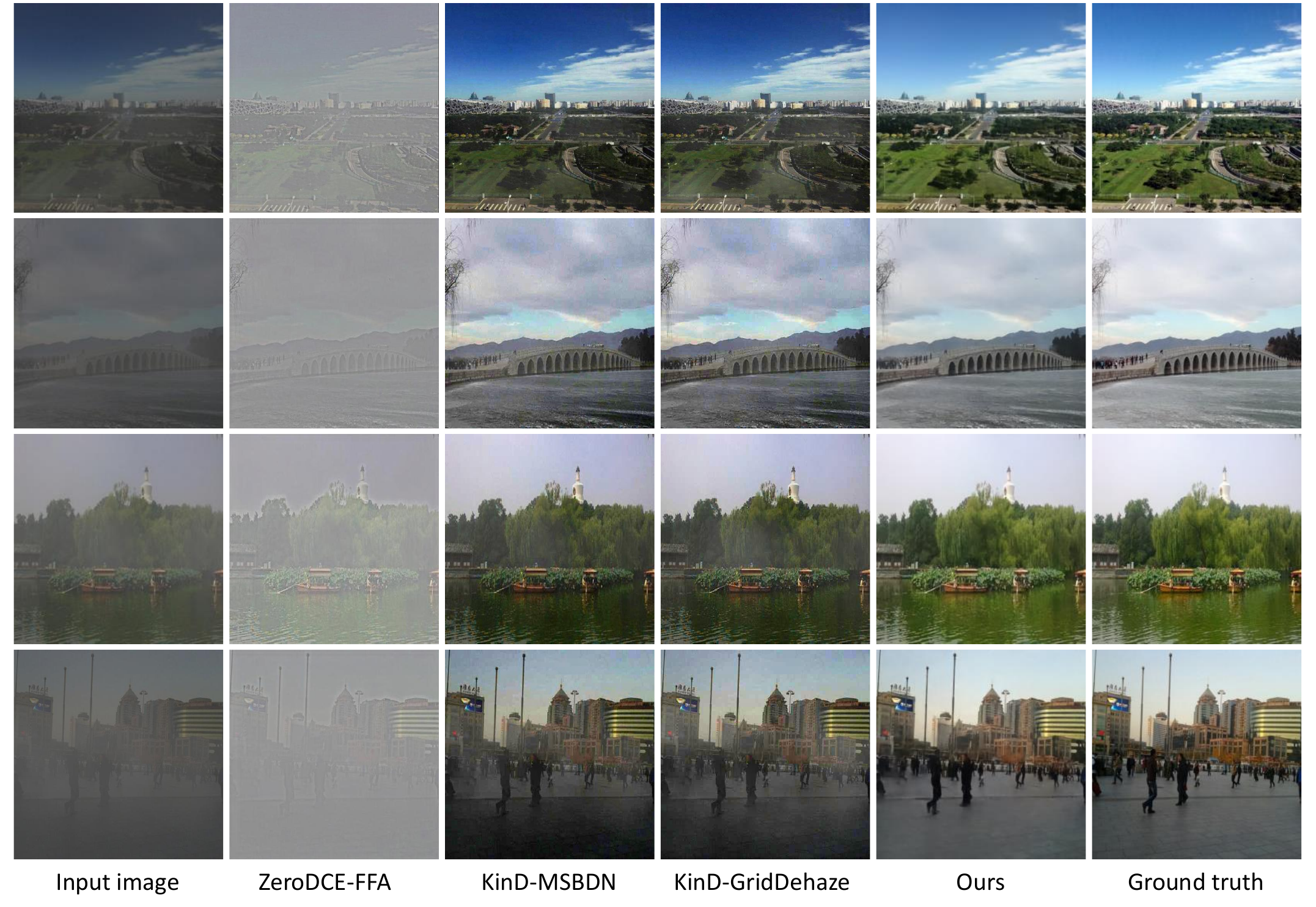} 
		\end{center}
		\caption{Qualitative comparisons among the state-of-the-art \textsl{ $LLE \rightarrow Dehaze$} methods trained and tested on visibility enhancement dataset. Our method recovers a high-quality image, while others have varying degrees of haze residue.}
		\label{fig:cmp_enhance_dehaze}
		\vspace{-1mm}
	\end{figure*}
	
	\begin{table}[b]
		\begin{center}
			\setlength{\tabcolsep}{0.9mm}
			\caption{Quantitative comparison results of $LLE \rightarrow Dehaze$ methods.}
			\label{tab:numerical_enhance_dehaze}
			\small
			\begin{tabular}{l |ccc | ccc | c}
				\toprule[1.2pt] 
				First(Retrained)-  & \multicolumn{ 3}{|c|}{KinD} & \multicolumn{ 3}{c|}{ZeroDCE} & \multirow{2}{*}{Ours} \\
				Second(Original)- & FFA & MSBDN & GridDehaze & FFA & MSBDN & GridDehaze & \\
				
				\hline
				\specialrule{0em}{1pt}{1pt}
				SSIM        & 0.752 & 0.751 & 0.766 & 0.523 & 0.463 & 0.211 & \textbf{0.914} \\
				PSNR         & 18.21 & 17.85 & 18.38 & 11.14 & 11.03 & 7.13 & \textbf{26.92} \\
				\hline
            \hline
                First(Retrained)-  & \multicolumn{ 3}{|c|}{KinD} & \multicolumn{ 3}{c|}{ZeroDCE} & \multirow{2}{*}{Ours} \\
				Second(Finetune)- & FFA$\dag$ & MSBDN$\dag$ & GridDehaze$\dag$ & FFA$\dag$ & MSBDN$\dag$ & GridDehaze$\dag$ & \\
				
				\hline
				SSIM        & 0.781 & 0.835 & 0.833 & 0.572 & 0.554 & 0.518 & \textbf{0.914} \\
				PSNR         & 21.14 & 24.72 & 23.49 & 14.16 & 11.43 & 10.18 & \textbf{26.92} \\
				\bottomrule[1.2pt]
			\end{tabular}  
		\end{center}
	\end{table}
	
	\begin{figure*} [t]
		\begin{center}
			\includegraphics[width=\linewidth]{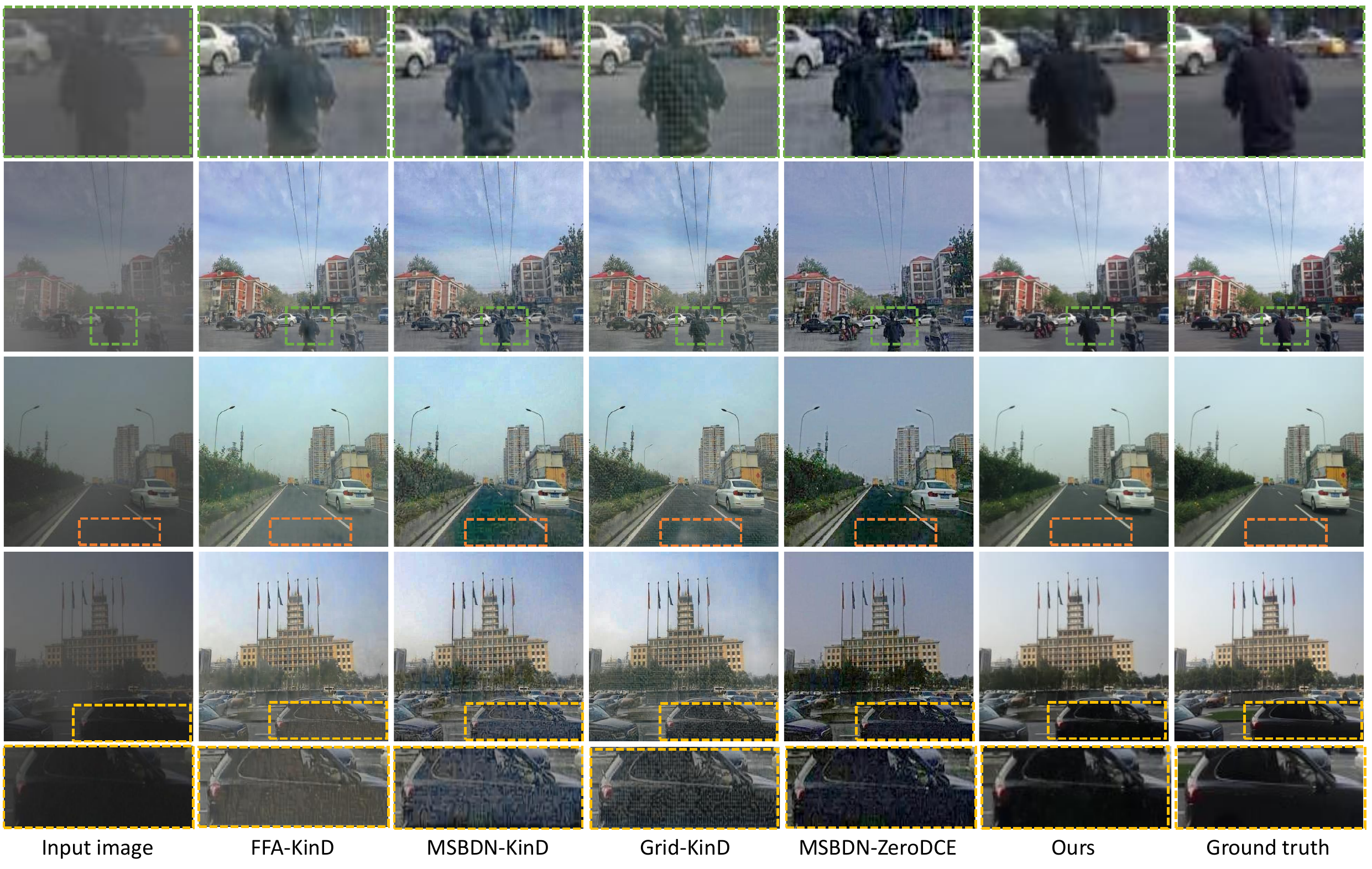}
		\end{center}
		\caption{Qualitative comparisons among the state-of-the-art \textsl{ $Dehaze \rightarrow LLE$} methods trained and tested on visibility enhancement dataset. Boxes indicate obvious differences. Nearly all haze is removed by our method without generating other noise while other methods are commonly not.}
		\label{fig:cmp_dehaze_enhance}
		\vspace{-3mm}
	\end{figure*}

	\begin{figure*}[t]
		\begin{center}
			\includegraphics[width=\linewidth]{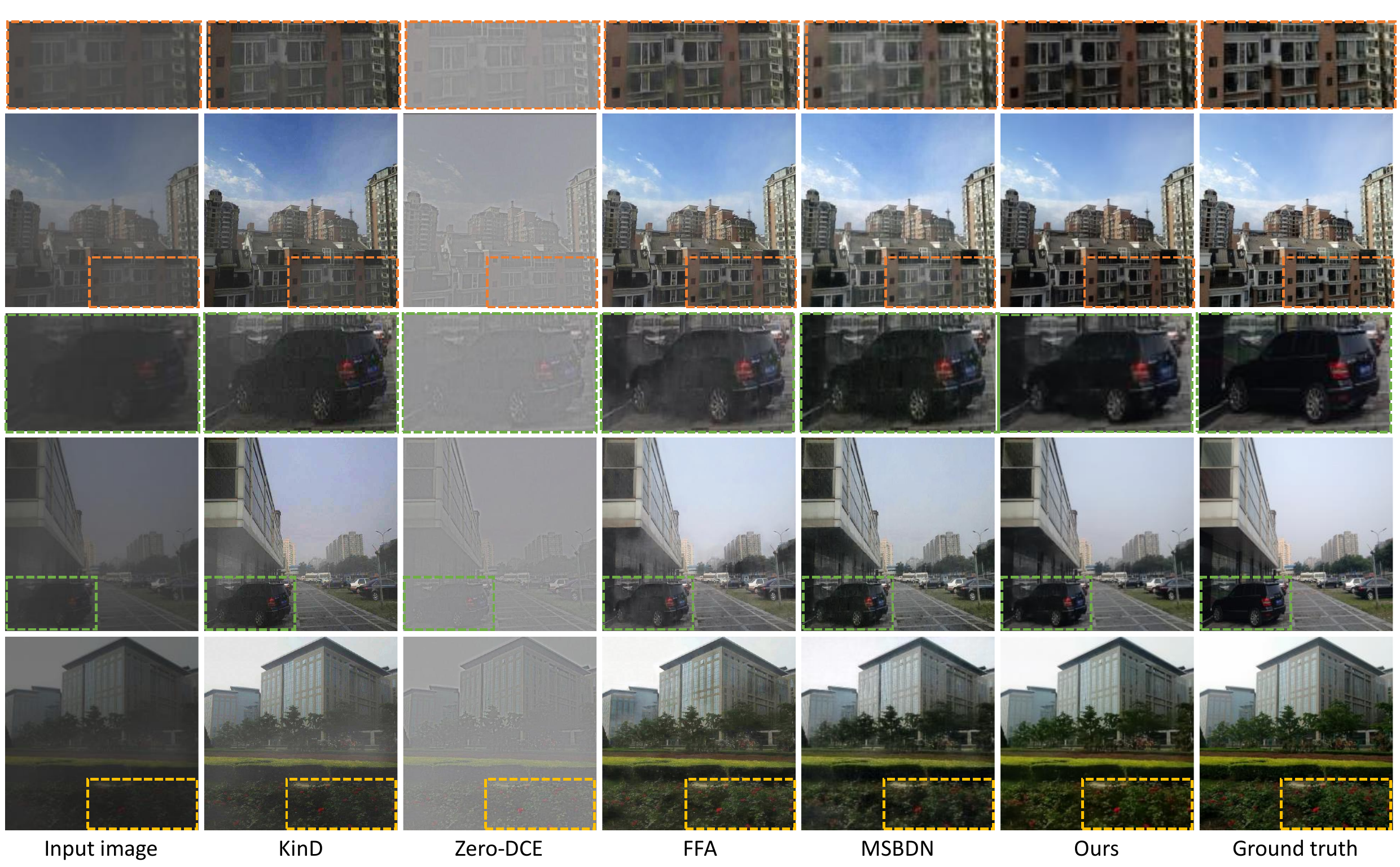} 
		\end{center}
		\caption{Qualitative comparisons among the state-of-the-art \textsl{end-to-end retrained} methods trained and tested on visibility enhancement dataset. Boxes indicate obvious differences. Our method can restore color details well and can remove the noise.}
		\label{fig:cmp_end_to_end}
		\vspace{-1mm}
	\end{figure*}
	\begin{table}[b]
        \vspace{-4mm}
	    \begin{center}
			\setlength{\tabcolsep}{0.5mm}
			\caption{Quantitative comparison results of $Dehaze \rightarrow LLE$ methods.}
			\label{tab:numerical_dehaze_enhance}
			\small
			\begin{tabular}{l | cc | cc | cc | c}
				\toprule[1.2pt]
				First(Retrained)-  & \multicolumn{ 2}{|c|}{FFA} & \multicolumn{ 2}{c|}{MSBDN} & \multicolumn{ 2}{c|}{GridDehaze} & \multirow{2}{*}{Ours} \\
				Second(Original)- & KinD & ZeroDCE & KinD & ZeroDCE & KinD & ZeroDCE & \\
				
				\hline
				\specialrule{0em}{1pt}{1pt}
				SSIM        & 0.776 & 0.650 & 0.744 & 0.614 & 0.751 & 0.612 & \textbf{0.914}\\
				PSNR         & 18.44 & 15.76 & 19.13 & 15.68 & 18.38 & 15.34 & \textbf{26.92}\\
				\hline
            \hline
                First(Retrained)-  & \multicolumn{ 2}{|c|}{FFA} & \multicolumn{ 2}{c|}{MSBDN} & \multicolumn{ 2}{c|}{GridDehaze} & \multirow{2}{*}{Ours} \\
				Second(Finetuned)- & KinD$\dag$ & ZeroDCE$\dag$ & KinD$\dag$ & ZeroDCE$\dag$ & KinD$\dag$ & ZeroDCE$\dag$ & \\
				
				\hline
                SSIM        & 0.629 & 0.678 & 0.633 & 0.654 & 0.672 & 0.641 & \textbf{0.914}\\
				PSNR         & 14.58 & 14.62 & 16.34 & 15.16 & 15.84 & 14.56 & \textbf{26.92}\\
				\bottomrule[1.2pt]
			\end{tabular}  
		\end{center}
	\end{table}
	\subsection{Implementation Details}
	The training examples are augmented by random rotating -10$^{\circ}\sim $10$^{\circ}$, and horizontal flipping. We implement the model using PyTorch and train it on an NVIDIA Tesla V100. We split the visibility enhancement dataset into two disjoint parts, 8073 images as the training set, and 897 images as the validation set. All the training images are resized to the size of 256 $\times$ 256. We train the framework for 250 epochs and optimize the parameters by Adam\cite{kingma2014adam} optimizer, where $\beta_1$ and $\beta_2$ take the values of 0.5 and 0.999. 
	\begin{figure*} [t]
        \begin{center}
        \includegraphics[width=\linewidth]{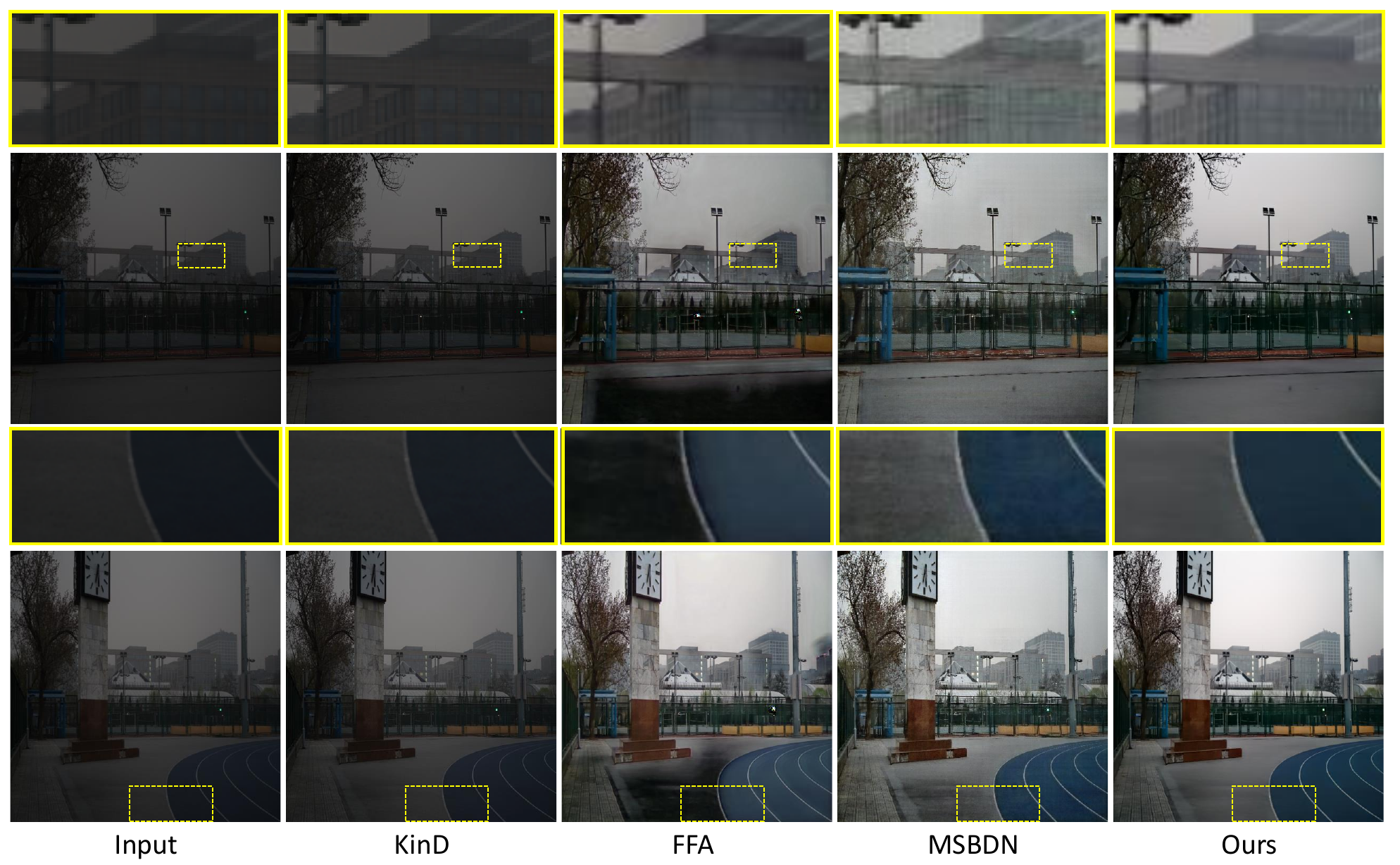}
        \end{center}
        \captionof{figure}{Examples of visibility enhancement results on our real dataset, compared with KinD~\cite{zhang2019kindling}, FFA~\cite{qin2020ffa}, MSBND~\cite{dong2020multi}.  Close-up views are displayed at the top of each image.}
        \label{fig:cmp_real}
    \end{figure*}
	\begin{algorithm*}[b]
        \caption{Pipeline to train our framework}
        \label{alg:pretrain_algorithm}
        \begin{algorithmic}
        \REQUIRE Datasets($I^H_L$, $I^H$, $I_L$, $I$), untrained block {$\mathcal{D}(\cdot, \theta)$, untrained block $\mathcal{E}(\cdot, \theta)$}, untrained CABF block
        
        \ENSURE Optimal $\mathcal{D}(\cdot, \theta)$, optimal $\mathcal{E}(\cdot, \theta)$ and optimal fushion block
        
        \enspace \textbf{Stage \uppercase\expandafter{\romannumeral1}} train for $\mathcal{D}(\cdot, \theta)$ and $\mathcal{E}(\cdot, \theta)$: 
        \end{algorithmic}
        \begin{algorithmic}[1] 
        \STATE train $\mathcal{D}(\cdot, \theta)$ with ($I^H_L$, $I_L$) and ($I^H$, $I$) using the dehazing loss
        \STATE train $\mathcal{E}(\cdot, \theta)$ with ($I^H_L$, $I^H$) and ($I_L$, $I$) using the enhancement loss
        
        \textbf{Stage \uppercase\expandafter{\romannumeral2}} jointly train for the cross-consistency framework: 
        \end{algorithmic}
        
        \begin{algorithmic}[1] 
        \STATE load pretrained blocks $\mathcal{D}(\cdot, \theta)$ and $\mathcal{E}(\cdot, \theta)$
        \STATE train the entire framework with ($I^H_L$, $I^H$, $I_L$, $I$) using the total loss
        \end{algorithmic}
    \end{algorithm*}

We first train the dehazing and enhancement blocks using corresponding data(hazy/clear images under different exposed conditions for dehazing block; lowlight/normal images under different concentrations of haze for enhancement block), and then we comprehensively train the entire framework, as shown in Algorithm~\ref{alg:pretrain_algorithm}. In the comprehensive training, we consider the noise problem of the second stage of each path, and strengthen the weight of the loss of the second part and the fusion part. We set $\lambda_{11} = \lambda_{21} = 0.2$, $\lambda_{12} = \lambda_{22} = 2$ and $\lambda_3 = 5$. The pre-trained blocks' learning rate is set as $10^{-3}$. In the comprehensive training, the two blocks' learning rate is set as $10^{-5}$ since them have been pre-trained while the fusion part's is set as $10^{-3}$.

	\subsection{Different Sequential Combination Manner}

	\textit{$LLE \rightarrow Dehaze$. } Recall that the visibility enhancement in this paper can be decomposed into two relative tasks, \ie, low-light enhancement and haze removal. 
	To evaluate the different combinations of the above tasks, we conduct the following experiments.
	Firstly, we use $LLE \rightarrow Dehaze$ manner to improve the low-visibility images.
	In detail, the first stage (low-light enhancement) enhances the low-light hazy image($I^H_L$) to normal hazy image($I^H$). 
	The current low-light enhancement methods are trained on the dataset without haze.
	To make such methods generalize well in our situation, we re-train the first stage methods on our simulated visibility enhancement dataset to make them learn the mapping $I^H_L \mapsto I^H$.
	We choose two recently proposed methods, KinD~\cite{zhang2019kindling} and ZeroDCE~\cite{Zero-DCE2020} as the first stage, respectively.
	After the first stage, the second stage becomes a traditional haze removal task. 
	Here we choose the SOTA dehazing methods (FFA/MSBDN/GridDehaze) as the second stage.
	We directly adopt their pre-trained model, which can be downloaded from their project page. 
	Furthermore, we finetune these methods on our dataset for a better mapping of $I^H \mapsto I$. The finetuned version of these methods is marked with $\dag$.

	We conduct the quantitative comparison on the test dataset of visibility enhancement dataset. Table~\ref{tab:numerical_enhance_dehaze} shows that our method obtains the highest values of metrics on both SSIM and PSNR. As mentioned above, the top part of the table is the result of directly using the pre-trained dehazing model, and the bottom part is using the fine-tuned models. Overall, the experimental results after fine-tuning are better than those directly using the pre-trained dehazing model. The visual comparisons are shown in Fig.~\ref{fig:cmp_enhance_dehaze}. As expected, the performance of these methods often suffers from color distortion and haze residual. The low-light enhancement will not only amplify the haze but corrupt the physical rule. Therefore, the dehazing methods cannot handle the further dehazing issue after the low-light enhancement.

	\textit{$Dehaze \rightarrow LLE$. }We alter the order of low-light enhancement and dehazing models to prove that performing image dehazing before low-light enhancement also suffers from unexpected results. 
    In the combined $Dehaze \rightarrow LLE$ methods, the one-stage dehazing method performs on the low-light hazy image($I^H_L$) and expects to obtain a low-light image without haze($I_L$). Since the models of traditional dehazing methods are trained under normal light conditions, which are not very suitable for our condition, we first retrain the dehazing methods (FFA-Net/MSBDN/GridDehaze) on our visibility enhancement dataset to make them learn the mapping $I^H_L \mapsto I_L$.
    The second stage is the traditional low-light enhancement task. So we first directly use the SOTA low-light enhancement methods' best open source pretrained models(KinD/Zero-DCE). Furthermore, we finetune these methods on our dataset for better mapping of $I_L \mapsto I$. The fine-tuned version of these methods is marked with $\dag$. 
	
	We conduct the quantitative comparison on the test dataset of the visibility enhancement dataset. Table~\ref{tab:numerical_dehaze_enhance} shows that our method receives the best numerical scores in both PSBR and SSIM. As mentioned above, the top part of the table is the result of directly using the pre-trained low-light enhancement model, and the bottom part is using the fine-tuned models. Since the former dehazing networks are not designed for the low-light hazy images, directly training them by visibility enhancement dataset cannot deliver pleasing results. Even though the enhancement models seem to perform well, the final dehazing results for low-light scenes still tend to leave haze and darken some regions. From the experimental results in the table and the visual results in Fig.~\ref{fig:cmp_dehaze_enhance}, it can be seen that the noise and artifacts generated by the residual haze in the first stage will be more severely amplified in the second stage. When the results obtained in the first stage are used to fine-tune the low light enhancement method, it will degrade the performance of the result.

	\subsection{End-to-end Manner}
	
	Logically, visibility enhancement for low-light hazy scenes could be included in both of dehazing task and low-light enhancement task. Therefore, with the proper dataset, both of these models may solve this task. To verify the effectiveness of proposed method, we directly retrain the SOTA methods of dehazing and low-light enhancement on our visibility enhancement dataset for comparison. 
	\begin{table}[b]
		\begin{center}
			\setlength{\tabcolsep}{1.0mm}
			\renewcommand\arraystretch{1.2}
			\caption{Quantitative comparisons among end-to-end retrained methods and ours.}
			\label{tab:numerical_end_to_end}
			\small
			\begin{tabular}{lc cc cc cc cc cc cc}
				\toprule[1.2pt]
				Method   & KinD & ZeroDCE & FFA & MSBDN & GridDehaze & Ours \\
				\hline
				\specialrule{0em}{1pt}{1pt}
				SSIM        & 0.751 & 0.521 & 0.820 & 0.837 & 0.817 & \textbf{0.914}\\
				PSNR         & 18.38 & 10.93 & 23.45 & 25.63 & 21.88 & \textbf{26.92}\\
				\bottomrule[1.2pt]
			\end{tabular}
		\end{center}
	\end{table}
	
	As shown in Fig.~\ref{fig:cmp_end_to_end}, directly training these former methods can deliver better results than the sequential combination way, which verify that applying the image dehazing and low-light enhancement in a cascade way may amplify the error. We find that the haze cannot be removed properly through low-light enhancement methods like KinD and ZeroDCE. Since the target of these methods aims to make the outputs brighter without focusing on haze removal, there are still haze residuals contained in the outputs of state-of-the-art dehazing methods. Based on the idea of cross-consistency, we jointly optimize dehazing and enhancement in a path-invariant manner. As a result, the proposed method can deliver clean and bright results. Table~\ref{tab:numerical_end_to_end} shows that our method is more effective and accurate for joint dehazing and enhancement in low-light hazy scenes.

    \begin{figure}[t]
        \centering\includegraphics[width=0.8\textwidth]{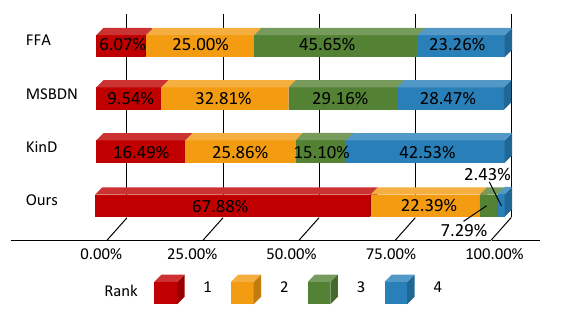}
        \centering\caption{Rating distribution of the user study. Results from 1 (best) to 4 (worst). Most participants rated our results as the highest(67.88\%) or second-highest(22.39\%) quality. Our method has significant advantages over other methods, suggesting that our results align with the high-quality standards of human perception.}
        \label{fig:user-study}
    \end{figure}

	\subsection{Performance on Real Scenarios}
	
	To evaluate the proposed method can generalize to real scenes, we also conduct comparative experiments in real scenes with four methods, which perform well on the simulated dataset. As shown in Fig.~\ref{fig:cmp_real}, our method can obviously deliver better results in terms of visual perception for this particular task. The performance on real scenarios also demonstrates the effectiveness of the proposed dataset simulation strategy. For real low-light hazy images, the conventional low-light enhancement methods cannot perform well. Though the dehazing methods can achieve good visual perception, the results still contain some artifacts and dark regions. 
	
	\textit{User study. } To further prove the effectiveness of our method, We invite 36 participants to attend a user study based on randomly selected 16 natural low-light hazy images. We apply three most competitive methods(KinD/FFA/MSBDN) and the proposed method on these cases for comparisons. For each case, the input data and the four results will be shown to the participants at the same time. The participants need to rank the quality of these four results from 1 (best) to 4 (worst) on the aspects of brightness, haze removal, artifacts, and noise. As shown in Fig.~\ref{fig:user-study}, the participants consider that the proposed method can deliver the best results in most of the cases, which shows that our results are more preferred by human subjects.
	\begin{figure}[t]
		\begin{center}
			\includegraphics[width=\linewidth]{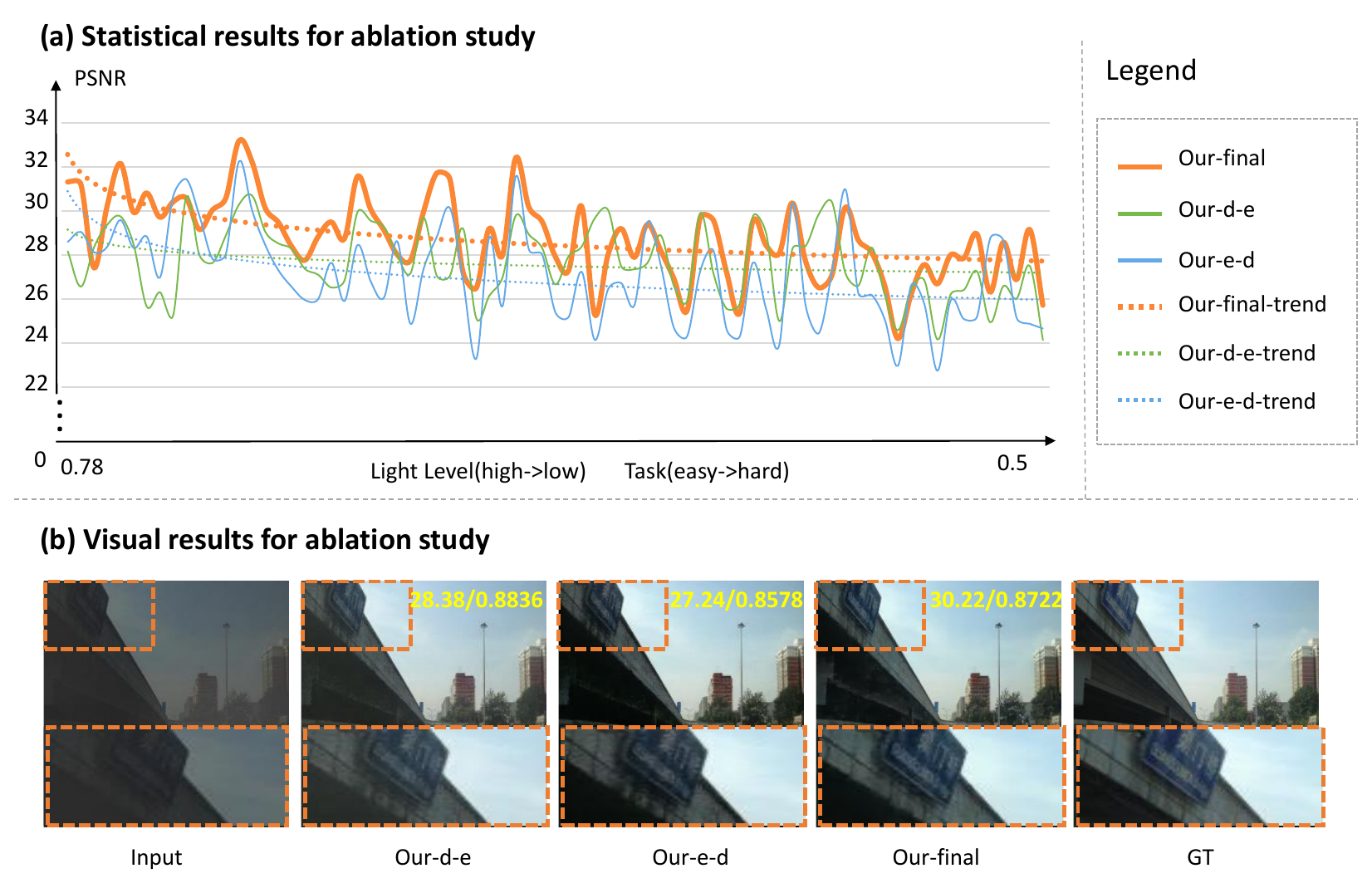} 
		\end{center}
		\caption{(a) Statistical results of ablation study. The legend shows the meaning of each line. The solid line is the original value. The dashed line is the trend fitted by Microsoft Excel. (b) Visual results of ablation study. Our final framework achieves favorable qualitative and quantitative results.} 
		\label{fig:ablation-study}
		\vspace{-5mm}
	\end{figure}
	
	 \begin{table}[b]
      \centering

      \caption{Ablation study.}
      \label{tab:ablation_study}
        \begin{tabular}{l |c  c  c  c}
            \toprule[1.2pt]
            Method & OnePath-d-e  & OnePath-e-d & TwoPath-without-fusion & Total model \\ 
            \hline
            \specialrule{0em}{1pt}{1pt}
            SSIM & 0.9096  & 0.9085 & 0.9116  & \textbf{0.9141}\\
            \hline
            PSNR & 26.40  & 26.47 & 26.46  & \textbf{26.92}\\
            
            \bottomrule[1.2pt]
        \end{tabular}
      \label{tab:ablation_asm}%
    \end{table}
	\subsection{Ablation Analysis}
    In this section, we analyze the effectiveness of our cross-consistency dehazing-enhancement framework. All pipelines are trained using the same setting. As mentioned in the proposed method, there are two paths to obtain the well-exposed clean image $I$ from the under-exposed hazy image $I^H_L$, as indicated in Eq.~\ref{eq:forward}. The experimental results proved that the order of dehazing and enhancement has a limited impact on the performance, as shown in Table~\ref{tab:ablation_study}. To perfectly utilize the results of two paths, we design the CABF block to obtain the final pleasing results. The results in the fourth row of Table~\ref{tab:ablation_study} demonstrate the effectiveness of the CABF block. In order to further analyze the existing numerical results, we count the results of the ablation experiments under different illumination difficulties. The results of our three methods are shown in the Fig.~\ref{fig:ablation-study}, and we used the Microsoft Excel to fit the performance of the three methods trend. It can be seen that as the illumination decreases, the task difficulty increases, and the performance of the three methods is in a downward trend as a whole. Still, the results of our final model always maintain the best performance results. At the same time, we selected a typical example for visual analysis. We can see that both the one path-d-e method and the one path-e-d method have different degrees of haze residue and other artifacts, and our final model can have a better visual effect. Our total model achieves the best results without increasing the number of parameters.
    Overall, the results of the ablation study verify the validity of the proposed cross-consistency dehazing-enhancement framework.

   \begin{figure}[t]
	\begin{center}
		\includegraphics[width=\linewidth]{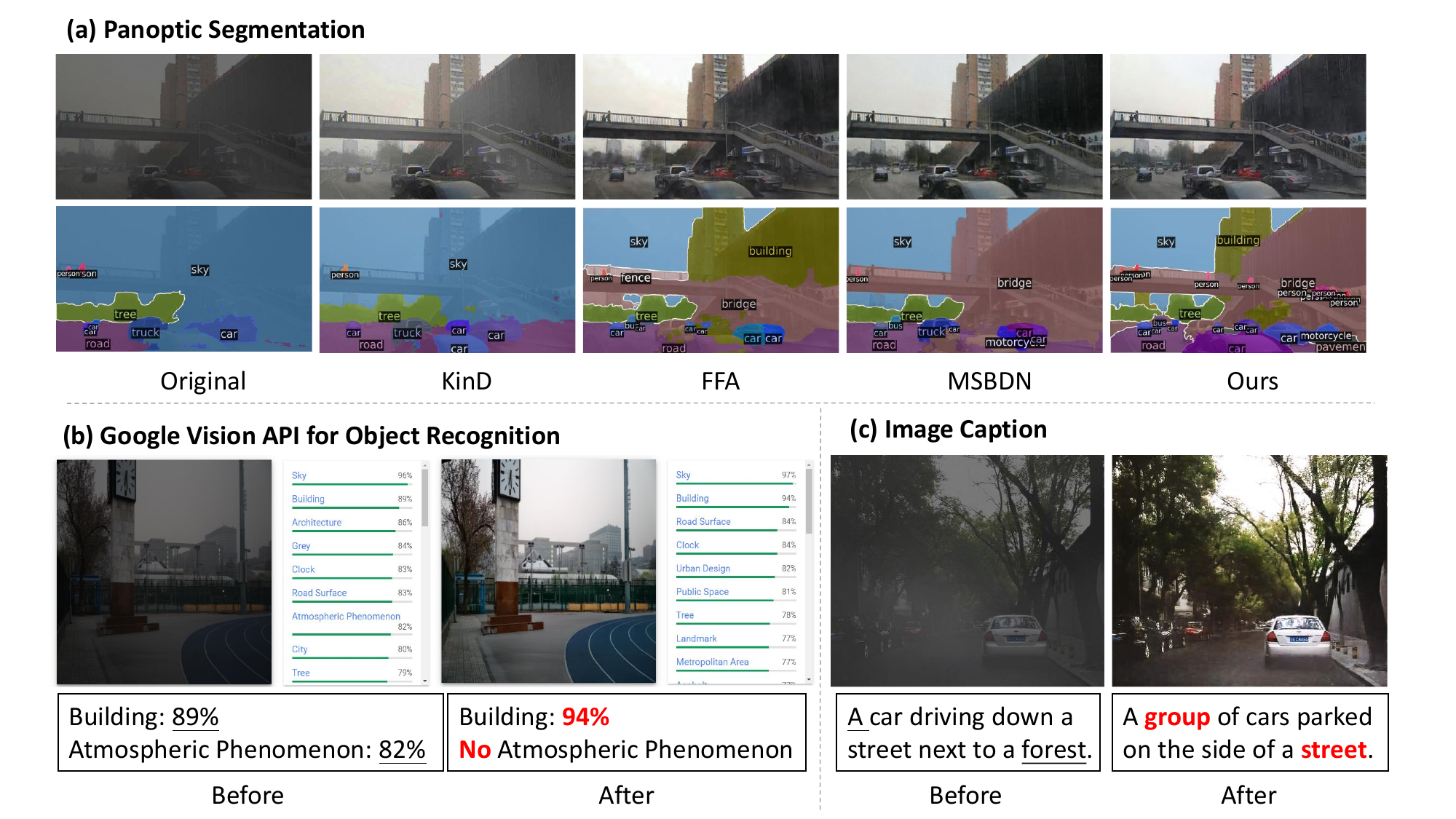} 
	\end{center}
	\caption{Results of three applications. (a) Visual comparisons on panoptic segmentation task with different enhancement and dehazing methods. (b) Google Vision API for Object Recognition before/after joint enhancement and dehaze. (c) Image caption before/after joint enhancement and dehaze. All three of these applications prove that our research is valuable.}
	\label{fig:application}
	\vspace{-3mm}
    \end{figure}

    \subsection{Applications}
    Since the hazy images with low-light illumination are inevitable in real scenarios, these images can become extremely hard cases for other computer vision tasks. To demonstrate the motivation and effectiveness of the proposed method, we conduct three computer vision applications: panoptic segmentation, object detection and image captioning. To verify the benefit of the our cross-consistency dehazing-enhancement framework, we directly apply our trained model as an image pre-processing tool. 
	
    For the panoptic segmentation task, we aim to pixel-wisely classify the category of instances in the image, including things in the foreground and background. We show the visual comparisons among different methods in Fig.~\ref{fig:application}. Due to the pleasing results of joint dehazing and enhancement for low-light scenes, our method can achieve desirable segmentation results for the input image.
    Image captioning task means  automatically generating a description of the image. In this case, we apply the self-critical~\cite{rennie2017self} to generate image captions. Fig.~\ref{fig:application} illustrates that the image captioning method generates more accurate descriptions when taking the image processed by our method as input. 
    We also employ Google Vision API to evaluate our results. As can be seen in Fig.~\ref{fig:application}, the Google API can recognize most objects in the results image rather than the original rainy image. Especially, the score of the building is improved by 5\% after visibility enhancement by our method. At the same time, the input image before processing will have a more serious atmospheric phenomenon.

	\section{Conclusions}
	 In this paper, we propose a cross-consistency dehazing-enhancement framework, focusing on integrating dehazing and low-light enhancement inference for low-light hazy scenes. With the parameter shared dehazing block and enhancement block, the proposed framework is capable of integrating the dehazing and enhancement without affecting each other. To tackle the problem of the lack of paired hazy low-light data, we propose an image simulation strategy to construct the visibility enhancement dataset. Leveraging on the dataset, we conduct several experimental comparisons to prove the necessity and effectiveness of our model. The experimental results demonstrate that the proposed solution can outperform the state-of-the-art methods for this particular task. In addition, the analysis of real scenes, user study, and applications show that the proposed method is practical and effective.

\bibliographystyle{unsrt}  
\bibliography{tomm}

\begin{thebibliography}{10}

\bibitem{narasimhan2002vision}
Srinivasa~G Narasimhan and Shree~K Nayar.
\newblock Vision and the atmosphere.
\newblock {\em International journal of computer vision}, 48(3):233--254, 2002.

\bibitem{wei2018deep}
Chen Wei, Wenjing Wang, Wenhan Yang, and Jiaying Liu.
\newblock Deep retinex decomposition for low-light enhancement.
\newblock {\em arXiv preprint arXiv:1808.04560}, 2018.

\bibitem{pizer1987adaptive}
Stephen~M Pizer, E~Philip Amburn, John~D Austin, Robert Cromartie, Ari
  Geselowitz, Trey Greer, Bart ter Haar~Romeny, John~B Zimmerman, and Karel
  Zuiderveld.
\newblock Adaptive histogram equalization and its variations.
\newblock {\em Computer vision, graphics, and image processing},
  39(3):355--368, 1987.

\bibitem{reza2004realization}
Ali~M Reza.
\newblock Realization of the contrast limited adaptive histogram equalization
  (clahe) for real-time image enhancement.
\newblock {\em Journal of VLSI signal processing systems for signal, image and
  video technology}, 38(1):35--44, 2004.

\bibitem{he2010DCP}
K.~{He}, J.~{Sun}, and X.~{Tang}.
\newblock Single image haze removal using dark channel prior.
\newblock {\em IEEE Transactions on Pattern Analysis and Machine Intelligence},
  33(12):2341--2353, 2011.

\bibitem{cai2016dehazenet}
Bolun Cai, Xiangmin Xu, Kui Jia, Chunmei Qing, and Dacheng Tao.
\newblock Dehazenet: An end-to-end system for single image haze removal.
\newblock {\em IEEE Transactions on Image Processing}, 25(11):5187--5198, 2016.

\bibitem{li2017all}
Boyi Li, Xiulian Peng, Zhangyang Wang, Jizheng Xu, and Dan Feng.
\newblock An all-in-one network for dehazing and beyond.
\newblock {\em arXiv preprint arXiv:1707.06543}, 2017.

\bibitem{zhang2018densely}
He~Zhang and Vishal~M Patel.
\newblock Densely connected pyramid dehazing network.
\newblock In {\em Proceedings of the IEEE conference on computer vision and
  pattern recognition}, pages 3194--3203, 2018.

\bibitem{liang2021fast}
Wei Liang, Jing Long, Kuan-Ching Li, Jianbo Xu, Nanjun Ma, and Xia Lei.
\newblock A fast defogging image recognition algorithm based on bilateral
  hybrid filtering.
\newblock {\em ACM transactions on multimedia computing, communications, and
  applications (TOMM)}, 17(2):1--16, 2021.

\bibitem{qin2020ffa}
Xu~Qin, Zhilin Wang, Yuanchao Bai, Xiaodong Xie, and Huizhu Jia.
\newblock Ffa-net: Feature fusion attention network for single image dehazing.
\newblock In {\em The Thirty-Fourth {AAAI} Conference on Artificial
  Intelligence, {AAAI} 2020, The Thirty-Second Innovative Applications of
  Artificial Intelligence Conference, {IAAI} 2020, The Tenth {AAAI} Symposium
  on Educational Advances in Artificial Intelligence, {EAAI} 2020, New York,
  NY, USA, February 7-12, 2020}, pages 11908--11915. {AAAI} Press, 2020.

\bibitem{dong2020multi}
Hang Dong, Jinshan Pan, Lei Xiang, Zhe Hu, Xinyi Zhang, Fei Wang, and
  Ming-Hsuan Yang.
\newblock Multi-scale boosted dehazing network with dense feature fusion.
\newblock In {\em Proceedings of the IEEE/CVF Conference on Computer Vision and
  Pattern Recognition}, pages 2157--2167, 2020.

\bibitem{Shao_2020_CVPR_DA}
Yuanjie Shao, Lerenhan Li, Wenqi Ren, Changxin Gao, and Nong Sang.
\newblock Domain adaptation for image dehazing.
\newblock In {\em Proceedings of the IEEE/CVF Conference on Computer Vision and
  Pattern Recognition (CVPR)}, June 2020.

\bibitem{liu2021syntheticMM}
Ye~Liu, Lei Zhu, Shunda Pei, Huazhu Fu, Jing Qin, Qing Zhang, Liang Wan, and
  Wei Feng.
\newblock From synthetic to real: Image dehazing collaborating with unlabeled
  real data.
\newblock {\em arXiv preprint arXiv:2108.02934}, 2021.

\bibitem{chen2021psdcvpr}
Zeyuan Chen, Yangchao Wang, Yang Yang, and Dong Liu.
\newblock Psd: Principled synthetic-to-real dehazing guided by physical priors.
\newblock In {\em Proceedings of the IEEE/CVF Conference on Computer Vision and
  Pattern Recognition}, pages 7180--7189, 2021.

\bibitem{zheng2021ultra}
Zhuoran Zheng, Wenqi Ren, Xiaochun Cao, Xiaobin Hu, Tao Wang, Fenglong Song,
  and Xiuyi Jia.
\newblock Ultra-high-definition image dehazing via multi-guided bilateral
  learning.
\newblock In {\em Proceedings of the IEEE/CVF Conference on Computer Vision and
  Pattern Recognition}, pages 16185--16194, 2021.

\bibitem{sun2022sadnet}
Ziyi Sun, Yunfeng Zhang, Fangxun Bao, Ping Wang, Xunxiang Yao, and Caiming
  Zhang.
\newblock Sadnet: semi-supervised single image dehazing method based on an
  attention mechanism.
\newblock {\em ACM Transactions on Multimedia Computing, Communications, and
  Applications (TOMM)}, 18(2):1--23, 2022.

\bibitem{du2021real}
Gaoming Du, Jiting Wu, Hongfang Cao, Kun Xing, Zhenmin Li, Duoli Zhang, and
  Xiaolei Wang.
\newblock A real-time effective fusion-based image defogging architecture on
  fpga.
\newblock {\em ACM Transactions on Multimedia Computing, Communications, and
  Applications (TOMM)}, 17(3):1--21, 2021.

\bibitem{huang2012gama_correct}
Shih-Chia Huang, Fan-Chieh Cheng, and Yi-Sheng Chiu.
\newblock Efficient contrast enhancement using adaptive gamma correction with
  weighting distribution.
\newblock {\em IEEE transactions on image processing}, 22(3):1032--1041, 2012.

\bibitem{hao2022decoupled}
Shijie Hao, Xu~Han, Yanrong Guo, and Meng Wang.
\newblock Decoupled low-light image enhancement.
\newblock {\em ACM Transactions on Multimedia Computing, Communications, and
  Applications (TOMM)}, 18(4):1--19, 2022.

\bibitem{shen2017MSR}
Liang Shen, Zihan Yue, Fan Feng, Quan Chen, Shihao Liu, and Jie Ma.
\newblock Msr-net: Low-light image enhancement using deep convolutional
  network.
\newblock {\em arXiv preprint arXiv:1711.02488}, 2017.

\bibitem{lv2018mbllen}
Feifan Lv, Feng Lu, Jianhua Wu, and Chongsoon Lim.
\newblock Mbllen: Low-light image/video enhancement using cnns.
\newblock In {\em BMVC}, page 220, 2018.

\bibitem{zhang2019kindling}
Yonghua Zhang, Jiawan Zhang, and Xiaojie Guo.
\newblock Kindling the darkness: A practical low-light image enhancer.
\newblock In {\em Proceedings of the 27th ACM International Conference on
  Multimedia}, MM '19, pages 1632--1640, New York, NY, USA, 2019. ACM.

\bibitem{Zero-DCE2020}
Chunle~Guo Guo, Chongyi Li, Jichang Guo, Chen~Change Loy, Junhui Hou, Sam
  Kwong, and Runmin Cong.
\newblock Zero-reference deep curve estimation for low-light image enhancement.
\newblock In {\em Proceedings of the IEEE conference on computer vision and
  pattern recognition (CVPR)}, pages 1780--1789, June 2020.

\bibitem{lamba2021restoring}
Mohit Lamba and Kaushik Mitra.
\newblock Restoring extremely dark images in real time.
\newblock In {\em Proceedings of the IEEE/CVF Conference on Computer Vision and
  Pattern Recognition}, pages 3487--3497, 2021.

\bibitem{zhang2014nighttime}
Jing Zhang, Yang Cao, and Zengfu Wang.
\newblock Nighttime haze removal based on a new imaging model.
\newblock In {\em 2014 IEEE International Conference on Image Processing
  (ICIP)}, pages 4557--4561. IEEE, 2014.

\bibitem{li2015nighttime}
Yu~Li, Robby~T Tan, and Michael~S Brown.
\newblock Nighttime haze removal with glow and multiple light colors.
\newblock In {\em Proceedings of the IEEE international conference on computer
  vision}, pages 226--234, 2015.

\bibitem{zhang2017fast}
Jing Zhang, Yang Cao, Shuai Fang, Yu~Kang, and Chang Wen~Chen.
\newblock Fast haze removal for nighttime image using maximum reflectance
  prior.
\newblock In {\em Proceedings of the IEEE conference on computer vision and
  pattern recognition}, pages 7418--7426, 2017.

\bibitem{zhang2020nighttime}
Jing Zhang, Yang Cao, Zheng-Jun Zha, and Dacheng Tao.
\newblock Nighttime dehazing with a synthetic benchmark.
\newblock In {\em Proceedings of the 28th ACM International Conference on
  Multimedia}, pages 2355--2363, 2020.

\bibitem{he2016deep}
Kaiming He, Xiangyu Zhang, Shaoqing Ren, and Jian Sun.
\newblock Deep residual learning for image recognition.
\newblock In {\em CVPR}, 2016.

\bibitem{zhang2018image}
Yulun Zhang, Kunpeng Li, Kai Li, Lichen Wang, Bineng Zhong, and Yun Fu.
\newblock Image super-resolution using very deep residual channel attention
  networks.
\newblock In {\em Proceedings of the European conference on computer vision
  (ECCV)}, pages 286--301, 2018.

\bibitem{wei2019single}
Kaixuan Wei, Jiaolong Yang, Ying Fu, Wipf David, and Hua Huang.
\newblock Single image reflection removal exploiting misaligned training data
  and network enhancements.
\newblock In {\em CVPR}, 2019.

\bibitem{liuICCV2019GridDehazeNet}
Xiaohong Liu, Yongrui Ma, Zhihao Shi, and Jun Chen.
\newblock Griddehazenet: Attention-based multi-scale network for image
  dehazing.
\newblock In {\em ICCV}, 2019.

\bibitem{mertens2007exposure}
Tom Mertens, Jan Kautz, and Frank Van~Reeth.
\newblock Exposure fusion.
\newblock In {\em 15th Pacific Conference on Computer Graphics and Applications
  (PG'07)}, pages 382--390. IEEE, 2007.

\bibitem{land1977retinex}
Edwin~H Land.
\newblock The retinex theory of color vision.
\newblock {\em Scientific american}, 237(6):108--129, 1977.

\bibitem{lore2017llnet}
Kin~Gwn Lore, Adedotun Akintayo, and Soumik Sarkar.
\newblock Llnet: A deep autoencoder approach to natural low-light image
  enhancement.
\newblock {\em Pattern Recognition}, 61:650--662, 2017.

\bibitem{lv2020fast}
Feifan Lv, Bo~Liu, and Feng Lu.
\newblock Fast enhancement for non-uniform illumination images using
  light-weight cnns, 2020.

\bibitem{li2017reside}
Boyi Li, Wenqi Ren, Dengpan Fu, Dacheng Tao, Dan Feng, Wenjun Zeng, and
  Zhangyang Wang.
\newblock Reside: A benchmark for single image dehazing.
\newblock {\em arXiv preprint arXiv:1712.04143}, 1, 2017.

\bibitem{Guo2019Cbdnet}
Shi Guo, Zifei Yan, Kai Zhang, Wangmeng Zuo, and Lei Zhang.
\newblock Toward convolutional blind denoising of real photographs.
\newblock {\em 2019 IEEE Conference on Computer Vision and Pattern Recognition
  (CVPR)}, 2019.

\bibitem{wang2004image}
Zhou Wang, Alan~C Bovik, Hamid~R Sheikh, Eero~P Simoncelli, et~al.
\newblock Image quality assessment: from error visibility to structural
  similarity.
\newblock {\em IEEE transactions on image processing}, 13(4):600--612, 2004.

\bibitem{kingma2014adam}
Diederik~P Kingma and Jimmy Ba.
\newblock Adam: A method for stochastic optimization.
\newblock {\em arXiv preprint arXiv:1412.6980}, 2014.

\bibitem{rennie2017self}
Steven~J Rennie, Etienne Marcheret, Youssef Mroueh, Jerret Ross, and Vaibhava
  Goel.
\newblock Self-critical sequence training for image captioning.
\newblock In {\em CVPR}, 2017.

\end{thebibliography}

\end{document}